\pdfoutput=1

\documentclass[11pt]{article}

\usepackage[]{acl}

\usepackage{hyperref}
\usepackage{times}
\usepackage{latexsym}

\usepackage[T1]{fontenc}

\usepackage[utf8]{inputenc}

\usepackage{microtype}

\usepackage{inconsolata}

\usepackage{listings} 
\usepackage{graphicx} 
\usepackage{float}
\usepackage{adjustbox}
\usepackage{booktabs} 
\usepackage{amssymb} 
\usepackage{multirow}
\usepackage{stfloats}
\usepackage{svg}
\usepackage{tikz}
\usepackage{forest}
\newcommand*\circled[1]{\tikz[baseline=(char.base)]{
            \node[shape=circle,draw,inner sep=2pt] (char) {#1};}}

\title{Puzzle Solving using Reasoning of Large Language Models: A Survey}

\author{Panagiotis Giadikiaroglou, Maria Lymperaiou, Giorgos Filandrianos, Giorgos Stamou \\
        Artificial Intelligence and Learning Systems Laboratory \\
        School of Electrical and Computer Engineering \\
        National Technical University of Athens \\
        panosgiadi@gmail.com, \{marialymp, geofila\}@islab.ntua.gr, gstam@cs.ntua.gr}

\begin{document}
\maketitle
\begin{abstract}

Exploring the capabilities of Large Language Models (LLMs) in puzzle solving unveils critical insights into their potential and challenges in 
AI, marking a significant step towards understanding their applicability in complex reasoning tasks. This survey leverages a unique taxonomy—dividing puzzles into rule-based and rule-less categories—to critically assess LLMs through various methodologies, including prompting techniques, neuro-symbolic approaches, and fine-tuning. Through a critical review of relevant datasets and benchmarks, we assess LLMs' performance, identifying significant challenges in complex puzzle scenarios. Our findings highlight the disparity between LLM capabilities and human-like reasoning, particularly in those requiring advanced logical inference. The survey underscores the necessity for novel strategies and richer datasets to advance LLMs' puzzle-solving proficiency and contribute to AI's logical reasoning and creative problem-solving advancements.
\end{abstract}

\section{Introduction}


Recent developments in 
LLMs such as GPT-3 \cite{Brown2020LanguageMA} and GPT-4 \cite{openai2023gpt4} have showcased their logical reasoning abilities across various domains \cite{Liu2023EvaluatingTL, Liu2023GLoREEL, Bao2023ASE, Creswell2022SelectionInferenceEL}. Despite these advances and their demonstrated capabilities in deductive reasoning \cite{Saparov2023TestingTG}, LLMs face limitations in inductive reasoning settings, as analyzed by \citet{xu2023large, Bang2023AMM}. The specific application of LLMs to puzzle solving, 
has not been thoroughly summarized.


\begin{figure}
\centering
\vskip -0.3cm
\includegraphics[width=0.9\linewidth]{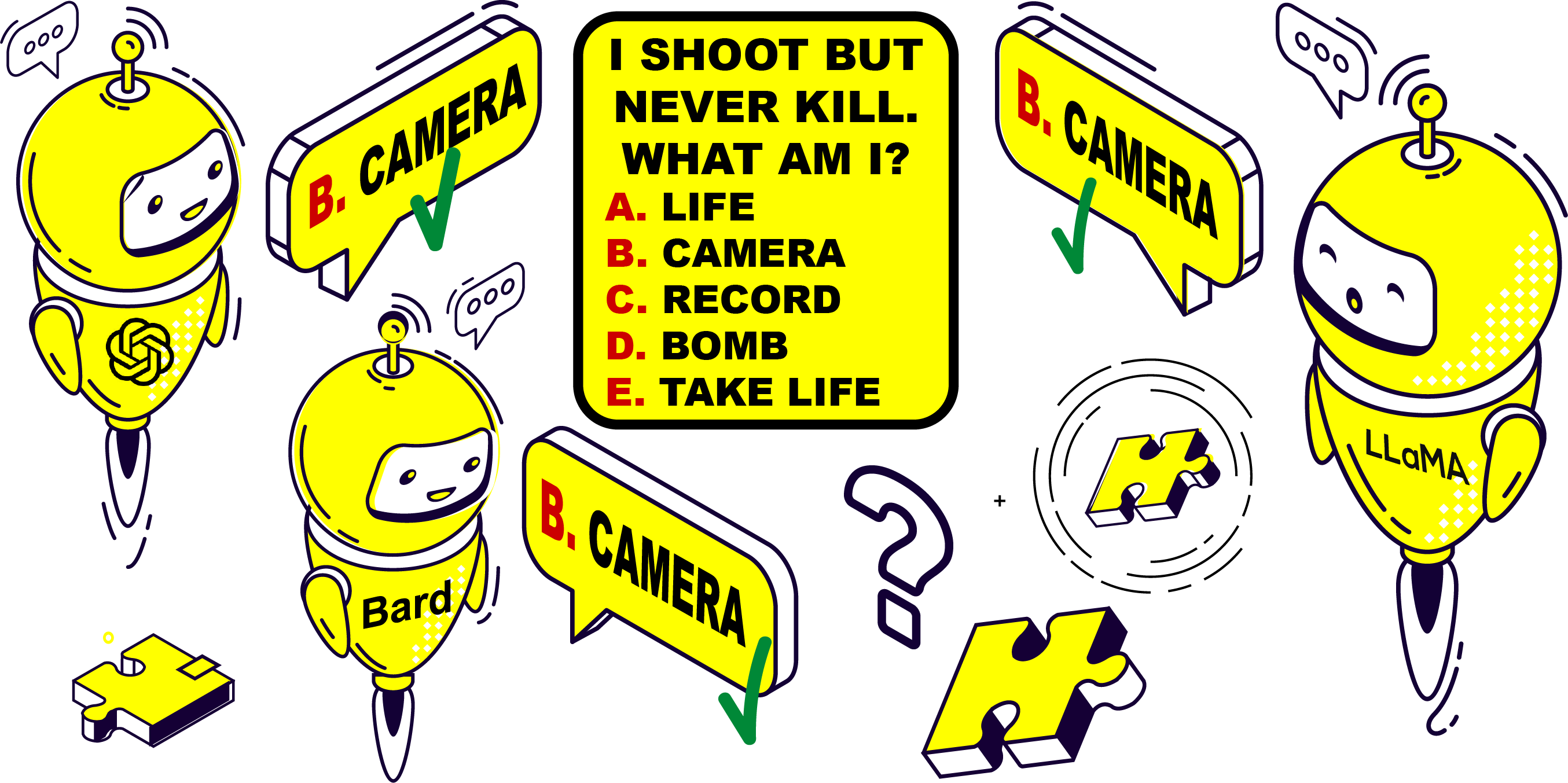} 
\caption{Riddle from RiddleSense \cite{Lin2021RiddleSenseRA}. GPT-4, LLaMA2-70B and Bard chose the right answer.}
\label{fig:puzzles}
\end{figure}



Our main contributions are as follows:
\circled{1} We introduce a distinction between rule-based and rule-less puzzles (§\ref{sec:taxonomy}), highlighting the varied knowledge demands necessary to tackle them.
\circled{2} We analyze the methodologies LLMs use to solve puzzles (§\ref{sec:methods}), assessing their impact on each category and comparing them with conventional problem-solving methods.
\circled{3} A detailed exploration of existing benchmarks that gauge models' reasoning abilities is conducted (§\ref{sec:datasets}).
\circled{4} Finally, this paper offers a detailed view of the present obstacles faced in puzzle-solving with LLMs and highlights a wide array of prospects for future research (§\ref{sec:discussion}).

Our categorization diverges from existing logical reasoning taxonomies by emphasizing on the underlying cognitive processes and the skills required for puzzle solving, rather than the question format \cite{Luo2023TowardsLA} or the nature of reasoning (deductive, inductive, abductive) \cite{Luo2023TowardsLA, yu2023natural, Yang2023LogicalRO, Qiao2022ReasoningWL, Huang2022TowardsRI, Flach2000AbductiveAI}. For instance, the existence of rules in puzzles such as Sudoku, Crosswords, or Minesweeper necessitates additional skills (e.g. strategy development) to correctly understand the game's rules or the ability to correctly format the output. In contrast, rule-less puzzles, such as riddles (Figure \ref{fig:puzzles}), programming challenges, and commonsense reasoning problems, leverage the model's inherent knowledge for solution derivation.

In our work, we define puzzles as problems that test cognitive abilities including logical reasoning, spatial cognition, and creative thinking by requiring the solver to discern patterns, apply deduction, and combine insights from available information in order to arrive at the correct solution.
Notably, we exclude puzzles that cannot be expressed in text in any way, such as jigsaw puzzles \cite{Markaki2022JigsawPS}, or problems that require multimodal understanding abilities of LLMs \cite{chia2024puzzlevqa, ghosal2024language}. Mathematical puzzles are also excluded, as this area diligently covered by the recent work of \citet{liu2023mathematical}.

We keep track of the latest progress in the field of puzzle solving using LLM reasoning in our GitHub \href{https://puzzlellms.github.io/}{https://puzzlellms.github.io/}.



\begin{figure*}[t]
\centering
\hspace*{-0.3cm} 
\begin{forest}
for tree={
    font=\scriptsize,
    grow=east,
    parent anchor=east,
    child anchor=west,
    rounded corners,
    draw,
    top color=white,
    bottom color=yellow!20,
    edge path={
        \noexpand\path [draw, \forestoption{edge}] (!u.east) -| (.child anchor)\forestoption{edge label};
    },
    l sep=7mm,
    s sep=3mm,
    anchor=west,
    inner xsep=7pt,
    inner ysep=3pt,
    calign=center,
    calign child=(n_children()+1)/2,
    if level=0{
        align=center,
        parent anchor=east,
        child anchor=west,
        inner xsep=0pt,
        draw=none,
    }{},
    if level=1{
        inner ysep=2pt,
        calign=child edge,
        s sep+=5pt,
        l sep=20pt,
        draw=none,
    }{},
    if level=2{
        inner ysep=0.75pt,
        calign=child edge,
        s sep+=5pt,
        l sep=15pt,
        draw=none,
    }{},
    if level=3{
        tier=tier3,
        l sep+=5pt,
        draw=none,
    }{},
    where level=3{text width=17em, draw=none}{},
}
[
\rotatebox{90}{\small{\textbf{Puzzle Categories}}}
[\parbox{10em}{\textbf{Rule-less puzzles:} rely more on flexible thinking, real-world knowledge and inferential reasoning}
        [\parbox{15em}{\textbf{Commonsense reasoning puzzles:} require understanding real-world situations and making inferences based on implicit knowledge}
            [\parbox{27em}{LatEval \cite{Huang2023LatEvalAI}, True Detective \cite{Del2022TrueDA}, DetectBench \cite{Gu2023GoBT}, MARB \cite{Tong2023EliminatingRV}}]
        ]
        [\parbox{15em}{\textbf{Programming puzzles:} involve analyzing or modifying code snippets to achieve a specific goal}
            [\parbox{27em}{P3 \cite{Schuster2021ProgrammingP}, \cite{savelka2023large}}]
        ]
        [\parbox{15em}{\textbf{Riddles:} use wordplay and metaphors to conceal the answers, requiring abstract connections and lateral thinking}
            [\parbox{27em}{BrainTeaser \cite{Jiang2023BRAINTEASERLT}, RiddleSense \cite{Lin2021RiddleSenseRA}, BiRdQA \cite{Zhang2021BiRdQAAB}, CC-Riddle \cite{Xu2022CCRiddleAQ}, PUZZLEQA \cite{Zhao2023SolvingAG}, MARB \cite{Tong2023EliminatingRV}}]
        ]
    ]
    [\parbox{10em}{\textbf{Rule-based puzzles:} provide explicit victory conditions, legal move sets or state transition rules that the model must follow to solve the puzzle}
    [\parbox{15em}{\textbf{Stochastic games:} incorporate randomness or hidden information, resulting in different outcomes}
            [\parbox{27em}{Minesweeper \cite{Li2023AssessingLP}, BoardgameQA \cite{Kazemi2023BoardgameQAAD}, Card Games \cite{Huang2024PokerGPTAE, Gupta2023AreCA}, Social Deduction Games \cite{Wang2023AvalonsGO, Xu2023ExploringLL, Lan2023LLMBasedAS}}]
        ]
    [\parbox{15em}{\textbf{Deterministic games:} provide all the information needed to produce an outcome from a given starting state and set of actions}
        [\parbox{27em}{BoardgameQA \cite{Kazemi2023BoardgameQAAD}, Sudoku \cite{Noever2021PuzzleSW, Long2023LargeLM, Ishay2023LeveragingLL}, Rubik's Cube \cite{Noever2021PuzzleSW, Ding2023EverythingOT}, Maze \cite{Noever2021PuzzleSW}, Crossword \cite{Yao2023TreeOT, Rozner2021DecryptingCC, Efrat2021CryptoniteAC, Kulshreshtha2022DownAA}, 8-puzzle \cite{Ding2023EverythingOT}, Game of 24 \cite{Ding2023EverythingOT, Yao2023TreeOT}, Chess \cite{Ishay2023LeveragingLL, Feng2023ChessGPTBP}}]
        ]
    ]
]
\end{forest}
\caption{A taxonomy of Puzzle Categories with the corresponding Datasets.}
\label{fig:taxonomy}
\end{figure*}
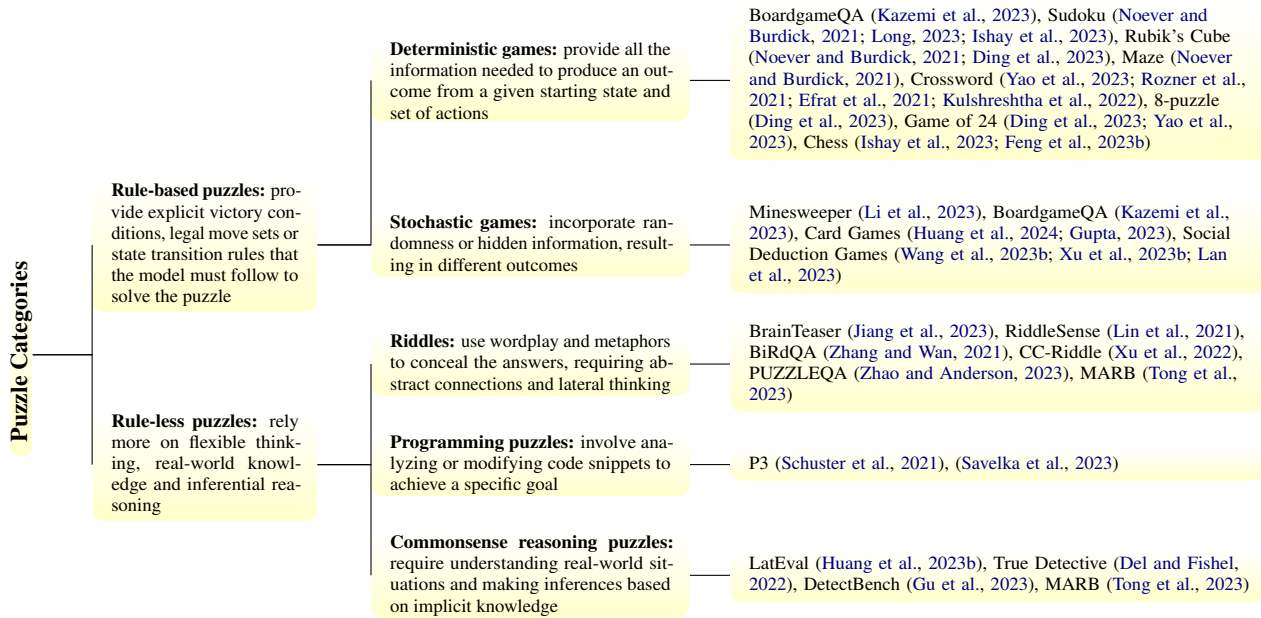


\section{Categorization of Puzzle Problems}\label{sec:taxonomy}

In assessing LLMs' reasoning capabilities, it is essential to categorize puzzles into coherent groups. We distinguish puzzles by their reliance on formal rules or broader world knowledge accompanied by general inferential skills, as illustrated in Figure \ref{fig:taxonomy}. This categorization not only highlights the cognitive diversity puzzles present, but also aligns with distinct reasoning challenges: rule-based puzzles demand logical deduction and strategic foresight within closed environments with defined parameters, whereas rule-less puzzles require general reasoning abilities, interpreting situations and explaining events by drawing inferences based on practical knowledge about the everyday world.


By separating puzzles into these categories, we aim to provide a nuanced analysis of LLMs' problem-solving abilities, reflecting on both structured challenges and those necessitating broader inferential reasoning.

\subsection{Rule-based Puzzles}
Rule-based Puzzles provide the model with explicit victory conditions, legal move sets or state transition rules. We further subdivide this category based on whether the state transitions are deterministic or incorporate randomness.

\textbf{Deterministic games}
always produce the same successor state given a current game state and action taken according to the rules. For example, in Chess, making a move always yields one unambiguous new board layout. Other examples include Sudoku, maze navigation, or solving a Rubik's cube. The model should learn strategies that operate within the possibility space defined by legal game mechanics.

\textbf{Stochastic games} incorporate randomness or hidden information, i.e. the same player action can lead to different probability distributions over next states. Examples include Minesweeper (hidden bomb locations) or card games e.g. Poker where opponents hold private hands. Mastering these games requires reasoning over uncertain states, planning multiple moves in advance and managing risk.

Thus, while both subgroups require logical reasoning bounded by formal rules, stochastic games pose the additional challenge of decision-making under uncertainty. Excelling in deterministic games enables pure reliance on deduction and forward search, while stochastic environments also require abilities for probabilistic inference, risk analysis, and reasoning with incomplete information. 

\subsection{Rule-less Puzzles}
Unlike rule-bounded puzzles, rule-less problems rely more on flexible thinking and 
real-world knowledge to interpret vague situations and infer unobserved details. Rather than testing systematic search or strategic planning, these puzzles measure cognitive skills for contextual interpretation, conceptual combination, and reasoning from common experiences. The following fall under this category.

\textbf{Riddles} utilize clever wordplay and literary devices to conceal answers. For example, "What gets wetter the more it dries?" obscures the solution of "a towel" through metaphor. Solving riddles requires making abstract connections between concepts hidden in lyrical language. This assesses skills for fluid reasoning, conceptual blending, and lateral thinking to decode linguistic relationships.

\textbf{Programming Puzzles} provide code snippets and require analyzing or modifying the underlying program logic. \citet{Schuster2021ProgrammingP} define a programming puzzle as a short Python program $f$, and the goal is to find an input which makes $f$ return True. Such puzzles assess skills like tracing execution, fixing errors, or anticipating outputs based on coding semantics. For example, the following puzzle tests understanding programming semantics to predict a system's behaviour: 
\begin{lstlisting}[language=Python, basicstyle=\small\ttfamily]
def mystery(x):
    return x // 2  
print(mystery(10))
\end{lstlisting}

\textbf{Commonsense Reasoning Puzzles} depict typical situations 
omitting key details. Solvers must explain events by inferring plausible implicit assumptions about motivations, causes and effects. For instance, the question "A man who was outside in the rain without an umbrella or hat didn’t get a single hair on his head wet. Why?" requires pragmatic analysis of unstated contextual factors.


\section{Methods and Strategies}\label{sec:methods}

In applying LLMs to puzzle solving, a wide array of methods and strategies enhances complex reasoning and performance. This section outlines the approaches used to address puzzles, aiming to highlight their application within this unique context. Given the extensive literature on prompt engineering and related methods \citet{besta2024demystifying, Chen2023UnleashingTP, Yu2023TowardsBC, Chu2023ASO, Qiao2022ReasoningWL, Liu2021PretrainPA}, we concentrate on the techniques most prevalent for puzzle solving, instead of describing each method separately. We divide existing methods into prompting techniques, neuro-symbolic approaches for puzzle translation and fine-tuning for specific domains. A detailed overview of the methods utilized across different puzzle categories is presented in Table \ref{table:methods}. We also discuss how conventional methods have faced these problems before the LLM era (App. \ref{appendix:conventional}).

\subsection{Prompting Methods}

Prompting strategies that provide intermediate reasoning steps are pivotal in enhancing the puzzle-solving capabilities of language models. The \textbf{few-shot in-context learning} paradigm offers one or more demonstrations within prompts, significantly improving performance for both rule-based and rule-less puzzles by showcasing the reasoning process without additional training \cite{Brown2020LanguageMA, dong2023survey, Zhou2022LargeLM}. 

Recent works focus on how different `thought structures' can guide LLMs to the final solution \cite{besta2024demystifying}.

\textbf{Chain topologies}, which include \textbf{Chain-of-Thought (CoT)} \cite{Wei2022ChainOT, Kojima2022LargeLM} have been applied to all kinds of puzzles, demonstrating their superiority over simple IO prompts. \textbf{Self-Refine} \cite{Madaan2023SelfRefineIR} is used for the Game of 24 (rule-based/deterministic), outperforming CoT with a 13\% higher success rate \cite{Yao2023TreeOT}. \citet{Gu2023GoBT} use several methods in a rule-less detective-style benchmark, including \textbf{Automatic CoT}, which autonomously generates diverse reasoning chains for various questions \cite{Zhang2022AutomaticCO}; \textbf{Complexity CoT} leverages the complexity of prompted chains, where more intricate reasoning steps often lead to improved performance in complex inference tasks by selecting outcomes that demonstrate deeper reasoning capabilities \cite{Fu2022ComplexityBasedPF}; and the \textbf{Plan-and-Solve (PS)} method, which uses two prompts for each problem—one to generate the reasoning process and the corresponding answer, and another to extract the final answer from the initial generation \cite{Wang2023PlanandSolvePI}. Despite the varied approaches, none of these methods clearly outperformed CoT across all tested LLMs. The best results are achieved by \textbf{Detective Thinking Prompt}, a CoT-like method introduced in the same study, which does not exceed the 61.6\% accuracy score of the best model, GPT-4. The method encourages the model to consider and analyze multiple clues within a given scenario, sequentially building towards a conclusion. This type of prompting can help the model handle complex scenarios where synthesizing disparate information correctly is crucial to generating accurate and logical outcomes.  \citet{Schuster2021ProgrammingP} exclusively utilized the solutions to programming puzzles that the model had already solved as examples, surpassing alternative approaches.

\textbf{Tree topologies} cover a variety of methods. \textbf{Self-Consistency (SC)} \cite{Wang2022SelfConsistencyIC} has been tested on rule-based/deterministic puzzles, such as the 8-puzzle, Game of 24 and Pocket Cube, as well as on rule-less commonsense reasoning puzzles, showcasing a small gain in the first category over CoT \cite{Ding2023EverythingOT, Yao2023TreeOT, Mo2023TreeOU} and no clear benefit in the second one \cite{Gu2023GoBT}. \textbf{Tree-of-Thought(s) (ToT)} \cite{Yao2023TreeOT, Long2023LargeLM} has been exclusively applied to rule-based/deterministic puzzles so far, achieving significantly improved success rates over CoT, with increases ranging from 26\% \cite{Mo2023TreeOU} to 70\% \cite{Yao2023TreeOT} depending on the puzzle and the depth of the tree, despite the increased LLM invocations \cite{Ding2023EverythingOT}. \textbf{Tree-of-Uncertain-Thought (TouT)} \cite{Mo2023TreeOU} achieved even better results than ToT on the same challenges, with a 9\% higher success rate on the Game of 24 and 3\% on mini-crosswords. Finally, \textbf{Inference-Exclusion-Prompting (IEP)} \cite{Tong2023EliminatingRV} employs a combination of forward and backward reasoning to approximate human logic and delivered some of the best results on riddles and commonsense puzzles when combined with CoT, scoring 82\% on puzzles--up from 81\% with zero-shot CoT--and 79\% on riddles, compared to 82\% with zero-shot CoT.

\textbf{Graph topologies} entail the following: \textbf{Graph-of-Thought(s) (GoT)} \cite{besta2023graph, Lei2023BoostingLR} and \textbf{Everything-of-Thought (XoT)} \cite{Ding2023EverythingOT} have been used to solve rule-based/deterministic puzzles. While GoT has shown poorer results compared to ToT, with a decrease ranging from 2\% to 6\% \cite{Ding2023EverythingOT}, XoT has been recognized as the most effective method for these puzzles. XoT integrates Monte Carlo Tree Search (MCTS) with LLMs for enhanced thought generation, achieving improvements in results from 53\% to 69\% compared to ToT. Additionally, XoT presents the fewest LLM invocations among the methods tested, including CoT, SC, ToT, and GoT.

A brief analysis of some basic methods not described here is presented in Appendix \ref{appendix:topologies}, while a more detailed analysis of all the methods discussed can be found in the extensive work of \citet{besta2024demystifying}. Beyond the aforementioned methods, the use of extra information such as \textbf{hints} for riddles and commonsense puzzles, or \textbf{introductions} and \textbf{summarizations} of the puzzles, has also been employed. The inclusion of supplementary details appears to yield positive results, although this is not always the case; for instance, Chinese riddles typically show worse results when hints are used \cite{Zhang2021BiRdQAAB}.

\subsection{Puzzle Translation}

In this subsection, we summarize the \textbf{neuro-symbolic techniques} used by LLMs to translate text puzzles from natural language into forms more amenable to solutions by external tools. Notably, these methods do not test the LLMs' puzzle solving capacity but rather assess their ability to encode puzzles into appropriate representations.

The primary approach involves using LLMs to generate \textbf{logic rules} from the puzzle's natural language and subsequently solve it using a symbolic solver. \citet{Ishay2023LeveragingLL} employ GPT-3 and GPT-4 to transform logic puzzles, such as chess puzzles, Jobs puzzle and Sudoku (rule-based/deterministic) into Answer Set Programming (ASP) formats by generating predicates and rules. They demonstrate that this method achieved significant results, with GPT-4 scoring 92\% accuracy in a logic puzzles dataset \citet{Mitra2015LearningTA}, compared to 7\% in few-shot and 21\% in zero-shot settings with the same model. They note that in few-shot settings, LLMs can generate complex programs that humans can easily refine and correct in case of code errors. Additionally, similar frameworks such as Logic-LM \cite{Pan2023LogicLMEL}, LINC \cite{Olausson2023LINCAN} and \citet{Yang2023NeuroSymbolicIB}'s method show promising results in logical reasoning tasks, although not specifically in puzzle settings.

While neuro-symbolic approaches have been applied to puzzle translation into logic rules, we have found no studies on transforming puzzles from natural language into \textbf{code}. However, techniques such as Program of Thoughts (PoT) prompting \cite{Chen2022ProgramOT} and Program-Aided Language (PAL) \cite{Gao2022PALPL} employ models to convert reasoning into Python programs for logical and mathematical reasoning datasets. Therefore, we encourage the research community to explore these methods for puzzle-solving tasks as well.

Given the structured nature of rule-based puzzles, this approach is inherently suitable for them. Consequently, it is logical that no studies have yet been conducted on rule-less puzzles in this context.

\subsection{Fine-Tuning}

Fine-tuning LLMs emerges as a potent strategy for enhancing their reasoning capabilities, ranging from general logical reasoning to specific puzzle-solving skills.

\subsubsection*{Logical Reasoning}

 LoGiPT \cite{Feng2023LanguageMC} is a language model fine-tuned to excel in logical reasoning tasks by mimicking the symbolic reasoning of logical solvers. The fine-tuning process involves constructing an instruction-tuning dataset comprising natural language (NL) logical questions paired with symbolic reasoning steps. It is fine-tuned to bypass syntax errors typically encountered in NL to symbolic language parsing, enabling it to directly produce answers.
 LogiT5 \cite{Luo2023TowardsLA} leverages a multi-task learning approach, incorporating diverse datasets to enhance its reasoning capabilities across different logical domains. The model is fine-tuned on the LOGIGLUE benchmark, which includes various logical reasoning datasets, enabling it to perform better on tasks with limited data by transferring knowledge across tasks. 
 
 \subsubsection*{Rule-based Puzzles}
 In the domain of rule-based deterministic puzzles, \citet{Noever2021PuzzleSW} observe suboptimal results when fine-tuning GPT-2 on Sudoku, Rubik's Cube and Mazes, potentially due to a brief fine-tuning period and limited training examples. Regarding crosswords, various studies \cite{Rozner2021DecryptingCC, Efrat2021CryptoniteAC} show mixed results, with some fine-tuned LLMs outperforming non-neural baselines and others not, highlighting the inherent challenge of cryptic crosswords for LLMs.  \citet{Kazemi2023BoardgameQAAD} demonstrate that fine-tuning LLMs with proofs and CoT under rule-based contexts yields some of the best results.

 \subsubsection*{Rule-less Puzzles}
 In the realm of riddles, the study of \citet{Lin2021RiddleSenseRA} illustrates that models like BERT \cite{devlin-etal-2019-bert}, RoBERTa \cite{Liu2019RoBERTaAR} and ALBERT \cite{Lan2019ALBERTAL} perform better when trained on both RiddleSense \citet{Lin2021RiddleSenseRA} and CommonsenseQA \cite{Talmor2019CommonsenseQAAQ} datasets, leveraging commonsense knowledge effectively. Moreover, \citet{Zhang2021BiRdQAAB} report that combining fine-tuning on ALBERT-XXL with transfer learning from CommonsenseQA achieved the highest accuracy, noting a 4\% improvement over simple fine-tuning. Lastly, the effectiveness of fine-tuning extends to commonsense reasoning \cite{Del2022TrueDA} and programming puzzles \cite{Schuster2021ProgrammingP}, showcasing its broad applicability across puzzle categories.

\section{Datasets, Benchmarks and Tasks}\label{sec:datasets}

Exploring diverse datasets, benchmarks, and tasks is crucial for evaluating LLMs in puzzle-solving. This section examines datasets within our puzzle taxonomy, encompassing formats, evaluation metrics, and methodologies. Figure \ref{fig:taxonomy} provides a detailed summary of datasets utilized across the taxonomy's categories, organized  according to puzzle type. The analysis demonstrates LLMs' versatility and the impact of techniques discussed in §\ref{sec:methods}. 

\subsection{Rule-based Puzzles}
We explore rule-based puzzles to assess LLMs' understanding within structured, closed-world environments. This includes deterministic puzzles such as Sudoku, Rubik's Cube, Crosswords, and the 8-puzzle, where solutions follow a set of defined rules. In contrast, stochastic games e.g. Minesweeper, card and social deduction games present variable outcomes from the same actions due to hidden factors. Research predominantly focuses on deterministic puzzles, highlighting a gap in addressing stochastic puzzle uncertainties—a promising direction for future research.

\subsubsection{Deterministic Puzzles}
\textbf{Sudoku} serves as a prime benchmark for LLMs due to its logical complexity.  \citet{Noever2021PuzzleSW} fine-tune GPT-2 \cite{Radford2019LanguageMA} on 1M Sudoku games, experimenting with compact single-string format, with empty cells represented by "-", and posited that a matrix representation may enhance the model's learning efficacy.  \citet{Long2023LargeLM} 
uses nested lists for puzzle representation\footnote{e.g. [[3,*,*,2], [1,*,3,*],[*,1,*,3],[4,*,*,1]]}, finding the Tree-of-Thought (ToT) method most effective, especially for smaller puzzles. \citet{Ishay2023LeveragingLL} explore neuro-symbolic approaches across Sudoku, Jobs puzzles and logic puzzles, demonstrating that well-prompted LLMs can accurately generate answer set programming rules.

For \textbf{Rubik's Cube} and \textbf{Maze solvers}, \citet{Noever2021PuzzleSW} assess GPT-2's spatial reasoning using over 2,400 Rubik's Cube samples and 10K mazes. Despite limited fine-tuning and token constrains, GPT-2 successfully solved the Rubik's Cube in 1 out of 7 attempts, showing potential despite a high rate of valid though incorrect solutions. \citet{Ding2023EverythingOT} apply multiple methods such as CoT, Self-Consistency, and various Thoughts (ToT, GoT, XoT) on a 2×2×2 Rubik's Cube using GPT-3.5 and GPT-4. XoT with self-revision emerges as most accurate, significantly outperforming others with a 77.6\% success rate.

Exploring LLM versatility, \citet{Ding2023EverythingOT} evaluate the effectiveness of XoT on the spatial \textbf{8-Puzzle} and numerical \textbf{Game of 24}. The 8-Puzzle's goal configuration challenges are solved with a remarkable 93.2\% accuracy across 419 puzzles using XoT with revision, showcasing superior efficiency over few-shot prompting and CoT. This high accuracy, coupled with a reduced number of LLM invocations, underscores the efficiency and potential of XoT in complex puzzle-solving contexts.

As for \textbf{Crosswords}, \citet{Rozner2021DecryptingCC} and \citet{Efrat2021CryptoniteAC} fine-tune T5 models \cite{Raffel2019ExploringTL} on extensive datasets of individual cryptic clues, revealing T5's advantage over traditional methods and highlighting areas for improvement, particularly with quick clues and specified answer lengths. \citet{Kulshreshtha2022DownAA}'s comparison of BART \cite{Lewis2019BARTDS} and T5 indicate a sub-30\% accuracy for clue-answer tasks, with retrieval-augmented generation transformers surpassing fine-tuned LLMs. Additionally, \citet{Yao2023TreeOT} apply 5-shot prompting and ToT to GPT-4 on Crossword puzzles significantly improving performance by solving 4 out of 20 puzzles and achieving a 60\% word-level success rate.


\citet{Feng2023ChessGPTBP} fine-tune two models, "ChessGPT" and "ChessCLIP," using a collection of 3.2M \textbf{chess puzzles} from the Lichess dataset\footnote{\href{https://lichess.org/}{https://lichess.org/}}. Each puzzle in the dataset include annotations for its rating, theme, and solution.

At last, \citet{Kazemi2023BoardgameQAAD} unveil \textbf{BoardgameQA}, a dataset featuring multi-choice questions against a backdrop of contradictory facts and rules. Models should navigate through these complexities to provide free-text answers. Their evaluation reveals that fine-tuning BERT-large and T5-XXL with proofs emerges as the most effective method, contrary to few-shot prompting on PaLM with CoT. Moreover, the presence of extra or conflicting information decreases accuracy.

\subsubsection{Stochastic Puzzles}
The \textbf{BoardgameQA} benchmark \cite{Kazemi2023BoardgameQAAD} also explores scenarios with missing information, which fall under the stochastic puzzle category. It is shown that as missing information increases, the accuracy of fine-tuned models decreases. However, this heightened difficulty does not similarly impact the performance of prompt-tuned and few-shot learning methods, which is likely due to the larger models that were applied.


\textbf{Minesweeper}, known for its hidden information and unpredictability, exemplifies stochastic puzzles, requiring players to deduce mine locations from numerical clues, challenging spatial reasoning. \citet{Li2023AssessingLP} evaluated LLMs on Minesweeper, comparing table and coordinate representations. Even though GPT-3.5 displayed initial understanding, enhancements like few-shot prompting had minimal effects. Conversely, GPT-4 improved mine identification but struggled to complete boards, highlighting Minesweeper's role in evaluating LLMs' strategic thinking. Experiments favored the coordinate representation over the table format for aiding LLM comprehension.

\textbf{Card games}, notably Poker, exemplify stochastic puzzles where strategic skill is crucial. Simplified Poker variants require players to infer opponents' cards and calculate odds amidst hidden intentions. \citet{Gupta2023AreCA} found that in Poker's pre-flop round, ChatGPT and GPT-4 grasp advanced strategies but do not reach Game Theory Optimal (GTO) play. ChatGPT leans towards a conservative approach, while GPT-4 exhibits more aggressive gameplay. \citet{Huang2024PokerGPTAE} leverage a Reinforcement Learning-trained OPT-1.3B model on all Poker phases revealing superior outcomes in win rates and efficiency, ultimately showcasing LLMs' adeptness at complex strategies in stochastic settings. An agent that leverages GPT-4 \cite{Guo2023SuspicionAgentPI} also achieves significant results in various imperfect information card games.

\textbf{Social deduction games}, including Werewolf and Avalon, blend logical reasoning with complex social dynamics, making them part of the broader stochastic puzzle domain. Such games challenge players to deduce roles involving unpredictable human behavior. 
\citet{Xu2023ExploringLL} propose a Werewolf framework using LLMs without tuning, leveraging historical interactions for strategic decisions and showcasing the models' ability in this context.
Similarly, frameworks for Avalon \cite{Wang2023AvalonsGO, Lan2023LLMBasedAS} show how LLMs can navigate scenarios demanding social manipulation and deduction, underscoring LLMs' proficiency in managing the complex interplay of logic and social interaction inherent in such games.

\subsection{Rule-less Puzzles}
This subsection delves into the diverse datasets related to rule-less puzzles, a category that predominantly encompasses riddles, programming puzzles, and commonsense reasoning challenges. Notably, we specifically focus on puzzles in their traditional sense, thereby excluding code generation datasets, which represent a distinct task type. A majority of rule-less puzzles are structured in a multiple-choice question-answering (QA) format, offering a standardized approach for evaluating LLMs' inferential reasoning. Benchmarks deviating from this format are specially mentioned, providing a broader perspective on the variety of rule-less puzzle datasets and their implications for LLM performance.

\subsubsection{Riddles}\label{subsubsec:riddle_datasets}
\textbf{RiddleSense} \cite{Lin2021RiddleSenseRA} offers a collection of 5.7K vertical thinking riddles, testing pre-trained LMs such as BERT, RoBERTa, ALBERT, and text-to-text QA models including UnifiedQA \cite{Khashabi2020UnifiedQACF} and T5. Larger LMs generally demonstrate better performance, with UnifiedQA using T5-3B leading, yet struggling with metaphors and counterfactual situations.

Complementing this, \textbf{BrainTeaser} \cite{Jiang2023BRAINTEASERLT} introduces 1119 lateral thinking puzzles. It contrasts instruction-based models (ChatGPT, T0, and FlanT5 \cite{Chung2022ScalingIL}) with commonsense ones (including RoBERTa variants and CAR \cite{Wang2023CARCR}). ChatGPT excels in both sentence-based and word-based puzzles, indicating its strength in lateral thinking. However, overall, LLMs still face challenges in exhibiting lateral thinking, with common errors in memorization and commonsense association. This dataset highlights the varied dimensions of reasoning that riddles can test, from vertical logic to lateral inference.

\textbf{BiRdQA} \cite{Zhang2021BiRdQAAB} explores the multilingual aspect of riddles, encompassing English and Chinese puzzles, while evaluating monolingual LMs (BERT, RoBERTa), as well as multilingual ones (mBERT, XLM-R \cite{Conneau2019UnsupervisedCR}). The use of brief riddle introductions and hints is also tested. Findings reveal a significant performance gap between LMs and human-level understanding, with monolingual models generally outperforming multilingual ones. Interestingly, additional context such as Wikipedia introductions and hints varied in effectiveness, with such aids benefiting English but not Chinese riddles.


\textbf{CC-Riddle} centers on 27K Chinese character riddles, involving multiple-choice, generative, and retrieval-based formats \cite{Xu2022CCRiddleAQ}. Evaluation demonstrates that models encountered difficulties in comprehension and exhibited misunderstandings, revealing the complexities inherent in character-based riddles.

In contrast, \textbf{PUZZLEQA} \cite{Zhao2023SolvingAG} offers 558 word puzzles in multiple choice and free-text formats. Larger models, e.g. GPT-3/3.5 show higher accuracy, especially in multiple-choice settings. However, methods such as CoT combined with summarization do not significantly enhance performance, pointing to the ongoing challenges in free-response puzzle solving.

Finally, \textbf{MARB} \cite{Tong2023EliminatingRV} encompasses a variety of riddle tasks. Several methodologies including zero-shot, CoT, IEP, and few-shot prompting are tested on models such as GPT-4 and PaLM2-540B \cite{Anil2023PaLM2T}. The combination of IEP and CoT emerged as the most effective method, highlighting the value of integrating multiple approaches for diverse riddle types. The dataset also includes commonsense puzzles (§\ref{subsubsec:commonsense_datasets}), showing similar trends with riddles.

\subsubsection{Programming Puzzles}
\textbf{P3 (Python Programming Puzzles)} \cite{Schuster2021ProgrammingP} offers a range of Python programming challenges, from straightforward string manipulations to complex tasks, such as the Tower of Hanoi and algorithmic puzzles, requiring from the model to find an input that makes the program $f$ return "True". Models applied to these puzzles include enumerative solvers for building Abstract Syntax Trees and autoregressive Language Model Solvers such as GPT-3 and Codex \cite{Chen2021EvaluatingLL}, employing varied prompting techniques. The evaluation metric pass@k, indicates the models' ability to solve a puzzle within a given number of attempts \cite{Chen2021EvaluatingLL}. Results show a correlation between puzzle difficulty for both models and humans, with descriptive prompts enhancing model performance. Interestingly, models proficient in code completion solved more puzzles with fewer tries, highlighting the importance of specialized capabilities in programming challenges.

\citet{savelka2023large} introduce a dataset comprised of 530 code snippets from programming courses, presenting puzzles in a multiple-choice format. The distinction between questions with and without code snippets offers a unique perspective on LLMs' problem-solving strategies. The dataset categorizes questions into six types, including true/false and output prediction. GPT models were evaluated, revealing that code inclusion significantly increases puzzle complexity. Accuracy rates vary, with higher performance on completion-oriented questions, suggesting that LLMs' effectiveness can depend heavily on question format and content.

While both P3 and Programming Snippets Dataset address programming puzzles, they do so in markedly different ways. P3's focus on finding correct Python program inputs contrasts with the multiple-choice format of the Programming Snippets Dataset. However, both datasets reveal key insights: descriptive prompts aid problem-solving, and question format significantly influences LLM performance.

\subsubsection{Commonsense Reasoning Puzzles}\label{subsubsec:commonsense_datasets}
True Detective \cite{Del2022TrueDA} presents detective puzzles in long-form stories, challenging LLMs such as GPT-3.5/4 to draw conclusions. Various methods, including CoT and Golden-CoT, are applied, revealing difficulties in making final inferences despite all necessary information being available. Golden-CoT provides the model with the reasoning behind the correct answer, so the model only needs to understand this reasoning and extract the answer. While Vanilla and CoT approaches perform close to random, Golden-CoT demonstrates significantly better accuracy, particularly with GPT-4. However, even with Golden-CoT, GPT-3.5 achieves a solve rate of only 63\%, whereas GPT-4 matches human solver results (without access to the reasoning behind the answer).

\textbf{DetectBench} \cite{Gu2023GoBT} containing 1200 questions, also evaluates informal reasoning in real-life contexts. It tests methods such as use of hints, various CoT approaches and detective thinking on models including GPT-4, GPT-3.5, GLM-4 and Llama2. Hints emerges as a powerful aid, with larger models generally outperforming smaller ones. The effectiveness of different approaches vary, with detective thinking effectively assisting most of the models.

Both datasets highlight the complexity of real-life reasoning and detective-style puzzles, demonstrating that hints play a crucial role in aiding both human and model performance.

\textbf{LatEval} \cite{Huang2023LatEvalAI} introduces a conversational format with English and Chinese stories, requiring players to ask yes/no questions before providing an answer. GPT-3.5, GPT-4, and various other Chat models are evaluated on their ability to ask relevant questions and maintain consistency with the truth. Larger models do not necessarily show advanced performance in question relevance. However, GPT-4 demonstrates the highest answer consistency, though there is still significant room for improvement. The dataset emphasizes the importance of interactive and conversational reasoning in commonsense understanding.

\textbf{PuzzTe} \cite{Szomiu2021APD}, with its array of comparison, knights and knaves, and zebra puzzles, represents a potentially rich resource for LLM testing. Despite not yet being applied to LLMs, its generated puzzle answers by Mace4 model finder and Prover9 theorem prover\footnote{\href{https://www.cs.unm.edu/~mccune/prover9/}{https://www.cs.unm.edu/~mccune/prover9/}} indicate its potential for future LLM evaluations.

The datasets under investigation demonstrate a variety of methods for evaluating commonsense reasoning in LLMs, ranging from detective-style puzzles to interactive story solving. Although larger models generally exhibit better performance, the complexity of these tasks poses significant challenges. Techniques such as sharing additional information through hints show effectiveness in improving outcomes, yet there remains a considerable gap between the performance of models and humans. It is important to note that in this work, we specifically focus on puzzle-oriented benchmarks, excluding general commonsense reasoning datasets e.g. CommonsenseQA, PIQA \cite{Bisk2019PIQARA} or StrategyQA \cite{Geva2021DidAU}.

\section{Discussion and Future Directions}\label{sec:discussion}

\textbf{Applied Methods and Dataset Gaps:}
Across our puzzle taxonomy, methods such as few-shot prompting, CoT, use of introductions and fine-tuning are commonly employed across most categories. Rule-based deterministic and rule-less commonsense puzzles show the greatest methodological variety, while riddles are also see diverse approaches. In contrast, rule-based stochastic and rule-less programming puzzles exhibit less variety, likely due to fewer studies in these areas. 
Figure \ref{fig:taxonomy} reveals that a substantial number of datasets are available for rule-based deterministic puzzles, such as Sudoku and Rubik's Cube, as well as a variety of rule-less riddles. This indicates a strong research interest and resource availability in these domains. However, there appears to be a scarcity of datasets for rule-based stochastic puzzles and rule-less programming puzzles. This gap highlights an opportunity for further research and dataset creation to introduce more diverse challenges for advancing the problem-solving capabilities of LLMs. The lack of benchmarks for stochastic puzzles led us to include tasks like card and social deduction games, which share core characteristics with traditional puzzles involving incomplete information. Additionally, neuro-symbolic techniques that translate natural language into code remain notably underutilized in puzzle benchmarks, suggesting a potential area for future exploration.

Comparatively, prompting methods like CoT and ToT enhance complex reasoning abilities without altering the underlying model parameters, yet their effectiveness on smaller LLMs requires further exploration. Moreover, methods such as ToT or GoT necessitate a greater number of model invocations compared to CoT and XoT, which could impact their efficiency and scalability \cite{Ding2023EverythingOT}. Fine-tuning, while fundamentally enhances reasoning by altering model parameters, is constrained by its specificity to particular tasks. For instance, models fine-tuned on CSQA have been observed to perform worst on the RiddleSense dataset than models fine-tuned directly on RiddleSense \cite{Lin2021RiddleSenseRA}, and a model fine-tuned on Poker games may not perform well on tasks like solving a Rubik’s cube. Finally, puzzle translation methods primarily test the model’s ability to interpret and rephrase problems rather than directly solve them, which tends to have a more limited impact on complex reasoning compared to the other two categories of methods.






\textbf{Performance Analysis:}

\textit{Rule-based / Deterministic}: Methods such as ToT and XoT (§ \ref{sec:methods}), typically enhance model reasoning abilities as the complexity of the structure increases \cite{Ding2023EverythingOT}. Yet, studies in BoardgameQA and crossword puzzles show generally poor model performance.

\noindent\textit{Rule-based/Stochastic}: Fine-tuning is prevalent here, enabling LLMs to grasp basic rules and simpler scenarios. However, they falter in complex settings that require extensive multi-step reasoning \cite{Li2023AssessingLP}.

\noindent\textit{Rule-less/Riddles \& Commonsense}: There is a notable performance gap between LLMs and human levels, with methods like CoT improving accuracy but still not matching human evaluation outcomes.

\noindent\textit{Rule-less/Programming}: LLMs find programming puzzles challenging, paralleling human difficulties \cite{Schuster2021ProgrammingP}. Tasks involving code analysis and reasoning in multiple-choice formats prove particularly tough \cite{savelka2023large}.

Furthermore, the format of questions significantly affects puzzle-solving effectiveness. Multiple-choice setups simplify tasks for LLMs by narrowing the solution search space, while free-text formats increase the difficulty level.

\textbf{Puzzle Generation} research is currently limited, likely because the ability to understand and solve puzzles is a prerequisite for generating them. In our survey, we primarily focused on puzzle-solving. The few works we found in puzzle generation reveal mixed results. For instance, GPT-3.5's attempts to generate puzzles with answers showed poor outcomes \cite{Zhao2023SolvingAG}. Conversely, the introduction of ACES, an autotelic generation method for diverse programming puzzles, demonstrates how semantic descriptors produced by LLMs can be leveraged for creative puzzle creation \cite{Pourcel2023ACESGD}. Lastly, there are recent works that have studied the generation of crossword puzzles of different languages, utilizing LLMs \cite{zugarini2024clueinstruct, Zeinalipour_2023, zeinalipour2023italian}.


\section{Conclusion}\label{sec:conclusion}

In this survey, we propose a taxonomy of puzzles for evaluating LLMs, categorizing them into rule-based (deterministic and stochastic) and rule-less puzzles (riddles, programming, and commonsense reasoning puzzles). We explore a spectrum of methods for LLM-based puzzle solving, ranging from prompting techniques to neuro-symbolic strategies and fine-tuning. By collating existing datasets in this domain, we provide a comprehensive overview of the resources available for such evaluations. Our analysis identifies current challenges, revealing a difficulty of most methods to successfully solve puzzles, while we outline future directions, emphasizing the need for advanced methodologies and diverse datasets to enhance LLMs' proficiency in puzzle solving.

\section*{Limitations}
In this study, we provide a survey of puzzle solving using reasoning of Large Language Models. Despite our best efforts, there may be still some limitations that remain in this paper. Firstly, due to the rapidly evolving nature of this field, we continuously add related approaches and analyses, but it is possible that some recent developments may not be included. Also, due to page constraints, we cannot extensively present all the methods nor provide all the technical details. This might limit the depth of understanding for some readers. Our review only includes methods within 4 years, primarily from sources such as ACL, EMNLP, NAACL, NeurIPS, ICLR, and arXiv. We plan to continue following these sources and adding new methods and datasets.
Additionally, all our conclusions §\ref{sec:conclusion} are based on empirical analysis. While this provides robust evidence, it may not capture all aspects of the problem. Lastly, as with any survey, our interpretations and conclusions §\ref{sec:discussion} are influenced by our own perspectives and understanding of the field. Other researchers might interpret the same studies differently. Despite these limitations, we believe this study provides a valuable overview of the current state of puzzle-solving using reasoning of Large Language Models.

\bibliography{anthology,acl_latex.bib}

\appendix

\section{Appendix}
\label{sec:appendix}

\subsection{Prompting Topologies}\label{appendix:topologies}

The \textbf{chain-of-thought (CoT)} paradigm involves step-wise explanatory reasoning chains, bolstering capabilities even in zero-shot settings with instructions such as "Let's think step-by-step" \cite{Wei2022ChainOT, Kojima2022LargeLM}. Complementing this, \textbf{self-consistency} generates multiple solution paths, selecting the most coherent one \cite{Wang2022SelfConsistencyIC}.

Exploring automated feedback, \citet{Pan2023AutomaticallyCL} examined \textbf{self-correction} within LLMs, noting its varied impact on logical reasoning. While instances of performance enhancement exist \cite{Weng2022LargeLM, Madaan2023SelfRefineIR}, broader gains are often elusive, with some strategies even detracting from overall reasoning accuracy \cite{Huang2023LargeLM}. However, \citet{Tyen2023LLMsCF} highlight the potential of backtracking methods, which, when informed about the specific location of errors, significantly boost the model’s correction abilities.

The \textbf{Tree-of-Uncertain-Thought (TouT)} prompting method structures problem-solving into a tree where each branch explores different uncertain reasoning pathways, allowing for multiple potential solutions \cite{Mo2023TreeOU}. In contrast, the \textbf{Tree-of-Thought(s)(ToT)} method \cite{Yao2023TreeOT, Long2023LargeLM} focuses on a more linear and deterministic approach, systematically breaking down problems into a single coherent pathway towards a solution. The \textbf{Graph-of-Thought(s) (GoT)} method \cite{besta2023graph, Lei2023BoostingLR} structures problem-solving by mapping out various interconnected reasoning pathways, allowing language models to explore and evaluate multiple solutions simultaneously within a flexible, network-like framework.

\subsection{Conventional Methods}\label{appendix:conventional}

AI and Machine Learning methods have long been applied to puzzles and games, with algorithms like Deep Blue \cite{CAMPBELL200257} and AlphaZero \cite{silver2017mastering} for Chess and Go, renowned for their exceptional results. This section contrasts ``traditional'' methods used to solve various puzzles with those derived from large language models (LLMs). Note that the aim of this paper isn't to determine the superior method for each puzzle, but to highlight the distinctive reasoning abilities of LLMs within diverse puzzle contexts. We particularly focus on rule-based puzzles, extensively addressed using conventional methods due to their structured, well-defined environments which require systematic strategies to achieve a solution. Conversely, rule-less puzzles such as riddles primarily test the logical, commonsense reasoning and creativity of models, without a clear path of steps to follow in order to find the solution, so we do not analyze this category.

\citet{chi2013techniques} utilized three techniques to solve \textbf{Sudoku}: backtracking, simulated annealing, and alternating projections. The backtracking method, a brute-force depth-first search, consistently resolves puzzles across all difficulty levels, albeit slowly. Constraint programming transforms Sudoku into a constraint satisfaction problem, swiftly enforcing constraints to deduce solutions, often within milliseconds \cite{Simonis2005SudokuAA}. These methods always find a solution for Sudoku puzzle, in contrast with LLMs that have not achieved results better than 80\% for 5x5 puzzles \cite{Long2023LargeLM}.

In their study on \textbf{Rubik's Cube}, \citet{Chen_2022} employed several traditional methods including Korf's algorithm \cite{Korf1997FindingOS}, which combines Iterative-Deepening Depth-First Search (IDDFS) with the A* algorithm and a heuristic search database. Both Thistlethwaite's \footnote{\href{https://www.jaapsch.net/puzzles/thistle.htm}{https://www.jaapsch.net/puzzles/thistle.htm}} and Kociemba's \footnote{\href{https://kociemba.org/}{https://kociemba.org/}} algorithms utilize group theory and similar search techniques to streamline the solving process, with Kociemba's version enhancing efficiency by simplifying the group structure. While all these algorithms effectively solve the Rubik's Cube—a task challenging for LLMs—Korf's method is particularly noted for its efficiency. Additionally, the study explored a machine learning strategy that integrates Monte-Carlo Tree Search (MCTS) with breadth-first search, yielding more optimized solutions, albeit at a lower efficiency. There have also been various attemts to solve Rubik's Cube using Reinforcement Learning (RL) like DeepCubeA \cite{mcaleer2018solving, Agostinelli2019SolvingTR} and others \cite{takano2023selfsupervision}, which although find a solution in relatively few steps are time-consuming, with duration varying from 38.7 to 75.6 seconds \cite{takano2023selfsupervision}.

\textbf{Mazes} are puzzles that can be solved by applying simple algorithms like depth-first search, A* or Trémaux's algorithm. However these problems are good for testing the spatial reasoning of LLMs. RL has also been utilized to solve mazes with \cite{barj2024reinforcement} leveraging LLM feedback during training.

In \citet{Ding2023EverythingOT} MCTS has been used to solve \textbf{Game of 24}, \textbf{8-Puzzle} and \textbf{Pocket Cube}, achieving surpassing many LLM techniques, including CoT, CoT-SC, ToT and GoT. Additionally, \citet{Rozner2021DecryptingCC} besides fine-tuning T5 for solving cryptic crosswords, have also used non-neural baselines including a WordNet-based heuristic model, a K-Nearest Neighbours bag of words model and a rule-based model, showing that the fine-tuning of T5 had the best results among them.

Finally, \citet{Studholme2001MinesweeperAA} proposed a method for solving \textbf{Minesweeper} by considering it as a constraint satisfaction problem (CSP). The core strategy involves transforming the game's challenges into a set of logical constraints that must be satisfied to avoid mines effectively.

In conclusion, most conventional methods used to solve rule-based puzzles employ deterministic approaches that reliably produce solutions, in stark contrast to the unpredictable nature of LLMs. Another advantage of these traditional methods is their explainability and interpretability, crucial attributes for thoroughly evaluating algorithms and understanding their decision-making processes. However, as demonstrated in the study by \citet{takano2023selfsupervision}, these methods can sometimes exhibit increased time complexity, indicating a potential trade-off between reliability and efficiency.

\subsection{Tables}
Table \ref{table:methods} delineates the various methods leveraged for puzzle-solving based on the datasets we have collected, illustrating the landscape of current LLM research in this domain. It particularly highlights the extensive methods applied to rule-based deterministic and rule-less commonsense puzzles. The absence of neuro-symbolic techniques and selection inference prompting indicates potential areas for expansion, especially considering their prospective benefits for LLMs grounded in logical reasoning datasets. 
\begin{table*}[bp]
\centering
\small 
\setlength{\tabcolsep}{4pt} 
\renewcommand{\arraystretch}{1.2} 
\begin{tabular}{|l|c|c|c|c|c|}
\hline
\textbf{Methods} & \multicolumn{2}{c|}{\textbf{Rule-based Puzzles}} & \multicolumn{3}{c|}{\textbf{Rule-less Puzzles}} \\ \cline{2-6} 
 & \textbf{Deterministic} & \textbf{Stochastic} & \textbf{Riddles} & \textbf{Programming} & \textbf{Commonsense} \\ \hline
\multicolumn{1}{|c|}{\textbf{Prompting}} & - & - & - & - & - \\ \hline
Few-shot & \checkmark & \checkmark & \checkmark & \checkmark & \checkmark \\ \hline
Chain-of-Thought & \checkmark & \checkmark & \checkmark & \checkmark & \checkmark \\ \hline
Self-refine & \checkmark &  &  &  &  \\ \hline
Auto-CoT &  &  &  &  & \checkmark \\ \hline
Complexity CoT &  &  &  &  & \checkmark \\ \hline
Plan \& Solve &  &  &  &  & \checkmark \\ \hline
Detective Thinking &  &  &  &  & \checkmark \\ \hline
Self-Consistency & \checkmark &  &  &  & \checkmark \\ \hline
Tree-of-Thoughts & \checkmark &  &  &  &  \\ \hline
Tree-of-uncertain-Thoughts & \checkmark &  &  &  &  \\ \hline
Inferential Exclusion Prompting &  &  & \checkmark &  & \checkmark \\ \hline
Graph-of-Thoughts & \checkmark &  &  &  &  \\ \hline
Everything-of-thoughts & \checkmark &  &  &  &  \\ \hline
Hints &  &  & \checkmark &  & \checkmark \\ \hline
Introduction/Summarization & \checkmark & \checkmark & \checkmark & \checkmark & \checkmark \\ \hline
\multicolumn{1}{|c|}{\textbf{Puzzle Translation}} & - & - & - & - & - \\ \hline
Logic & \checkmark &  &  &  & \\ \hline
Code &  &  &  &  &  \\ \hline
\multicolumn{1}{|c|}{\textbf{Fine-Tuning}} & \checkmark & \checkmark & \checkmark & \checkmark &  \checkmark \\ \hline
\end{tabular}
\caption{Methods used by each category of our taxonomy based on the puzzle benchmarks we collected}
\label{table:methods}
\end{table*}
\noindent The table further reflects the adaptability of certain methods like Chain-of-Thought, few-shot learning and fine-tuning, which are utilized across multiple puzzle types, hinting at their effectiveness. Based on this information, we not only catalogue the current state of method applications in puzzle-solving with LLMs but also highlight opportunities for innovative research in areas yet to be explored.

\vspace*{\fill}

\end{document}